\def\year{2020}\relax
\documentclass[letterpaper]{article} 
\usepackage{aaai20}  
\usepackage{times}  
\usepackage{helvet} 
\usepackage{courier}  
\usepackage[hyphens]{url}  
\usepackage[table]{xcolor}
\usepackage{color, colortbl}
\usepackage{adjustbox}
\usepackage{subcaption}
\usepackage{amsmath, amssymb}
\usepackage{hhline, multirow, makecell}
\usepackage{graphicx} 
\usepackage{graphicx}  
\title{RSM-GAN: A Convolutional Recurrent GAN for Anomaly Detection in \\ Contaminated Seasonal Multivariate Time Series}

\author{Farzaneh khoshnevisan \textsuperscript{\rm 1}, Zhewen Fan \textsuperscript{\rm 2} \\ \textsuperscript{1}{North Carolina State University}\\
\textsuperscript{2}{Intuit Inc.}\\
fkhoshn@ncsu.com,
zhewen\_fan@intuit.com
}

\begin{document}

\maketitle

\begin{abstract}
Robust anomaly detection is a requirement for monitoring complex modern systems with applications such as cyber-security, fraud prevention, and maintenance. These systems generate multiple correlated time series that are highly seasonal and noisy. This paper presents a novel unsupervised deep learning architecture for multivariate time series anomaly detection, called Robust Seasonal Multivariate Generative Adversarial Network (RSM-GAN). It extends recent advancements in GANs with adoption of convolutional-LSTM layers and an attention mechanism to produce state-of-the-art performance. We conduct extensive experiments to demonstrate the strength of our architecture in adjusting for complex seasonality patterns and handling severe levels of training data contamination. We also propose a novel anomaly score assignment and causal inference framework. We compare RSM-GAN with existing classical and deep-learning based anomaly detection models, and the results show that our architecture is associated with the lowest false positive rate and improves precision by 30\% and 16\% in real-world and synthetic data, respectively. Furthermore, we report the superiority of RSM-GAN regarding accurate root cause identification and NAB scores in all data settings.
\end{abstract}

\section{Introduction}
\noindent Detecting anomalies in real-time data sources is becoming increasingly important thanks to the steady rise in the complexity of modern systems. Examples of these systems are an AWS Cloudwatch service that tracks metrics such as CPU Utilization
and EC2 usage, or an enterprise data encryption process where multiple encryption keys coexist and are monitored. Anomaly detection (AD) applications include cyber-security, data quality maintenance, and fraud prevention. An effective AD algorithm needs to be accurate and timely to allow operators to take preventative and corrective measures before any catastrophic failure happens. Time series forecasting techniques such as Autoregressive Integrated Moving Average (ARIMA) \cite{arima} as well as Statistical Process Control (SPC) \cite{spc} were popular algorithms for such applications. However, a complex system often outputs multiple correlated information sources. These conventional AD techniques are not adequate to capture the inter-dependencies among multivariate time series (MTS) generated by the same system. As a result, many unsupervised density or distance-based models such as K-Nearest Neighbors \cite{knn} have been developed. However, these models usually ignore the temporal dependency and seasonality in time series. The importance of modeling temporal dependencies in time series has been well studied \cite{tsa}, and failing to capture them results in model mis-specification and a high false positive rate (FPR) \cite{ts1}. Seasonality is hard to model due to its irregular and complex nature. Most algorithms such as \cite{arima} make a simplistic assumption that there exists only one seasonal component such as a weekly or monthly seasonality, while in real-world complex systems, multiple seasonal patterns can occur simultaneously. Not accurately accounting for seasonality in AD can also lead to false detection \cite{ts2}. \\
\noindent Recent advancement in computation has afforded rapid development in deep learning-based AD techniques. Auto-encoder based models coupled with Recurrent Neural Network (RNN)
are well suited for capturing temporal and spatial dependencies and they detect anomalies by inspecting the reconstruction errors \cite{lstmed}. 
Generative Adversarial Networks (GANs) is another well-studied deep learning framework. The intuition behind using GANs for AD is to learn the data distribution, and in case of anomalies, the generator would fail to reconstruct input and produce large loss. GANs have enjoyed success in image AD  \cite{ganomaly,gan2,gan3image}, but have yet been applied to the MTS structure. Despite such advancements, none of the previous deep learning AD models addressed the seasonality problem. Furthermore, most deep learning model
rely on the assumption that the training data is normal with no contamination. However, real-world data generated by a complex system often contain noise or undetected anomalies (contamination).
Lastly, MTS anomaly detection task should not end at simply flagging the anomalous time points; a well-designed causal inference can help analysts narrow down the root cause(s) contributing to the irregularity, for them to take a more deliberate action. \\
\noindent To address the aforementioned problems in MTS AD, we propose an unsupervised adversarial learning architecture fully adopted for MTS anomaly detection tasks, called Robust Seasonal Multivariate Generative Adversarial Networks (RSM-GAN). Motivated by \cite{ganomaly,mscred}, we first convert the raw MTS input into multi-channel correlation matrices with image-like structure, and employ convolutional and recurrent neural network (Convolutional-LSTM) layers to capture the spatial and temporal dependencies. Simultaneous training of an additional encoder addresses the issue of training data contamination. While training the GAN, we exploit Wasserstein loss with gradient penalty \cite{wgan-gp} to achieve stable training and during our experiment it reduces the convergence time by half. Additionally, we propose a smoothed attention mechanism to model multiple seasonality patterns in MTS. In testing phase, residual correlation matrices along with our proposed scoring and causal inference framework are utilized for real-time anomaly detection. We conduct extensive empirical studies on synthetic datasets as well as an encryption key dataset. The results show superiority of RSM-GAN for timely and precise detection of anomalies over state-of-the-art baseline models. \\ 
\noindent The contributions of our work can be summarized as follows: (1) we propose a convolutional recurrent Wasserstein GAN architecture (RSM-GAN), and extend the scope of GAN-based AD from image to MTS tasks; (2) we model seasonality as part of the RSM-GAN architecture through a novel smoothed attention mechanism; (3) we apply an additional encoder to handle the contaminated training data; (4) we propose a scoring and causal inference framework to accurately and timely identify anomalies and to pinpoint unspecified numbers of root cause(s). The RSM-GAN framework enables system operators to react to abnormalities swiftly and in real-time manner, while giving them critical information about the root causes and severity of the anomalies.

\section{Related Work}

MTS anomaly detection has long been an active research area because of its critical importance in monitoring high risk tasks.
There are $3$ main types of detection methods: 1) classical time series analysis (TSA) based methods; 2) classical machine learning based methods; and 3) deep learning based methods. The TSA-based models include Vector Autoregression (VAR) \cite{var}, and latent state based models like Kalman Filters \cite{kalman}. These models are prone to mis-specification, and are sensitive to noisy training data. Classical machine learning methods can be further categorized into distance based methods such as the k-Nearest Neighbor (kNN) \cite{knn}, classification based methods such as One-Class SVM \cite{one-svm}, and ensemble methods such as Isolation Forest \cite{iforest}. These general purpose AD methods do not account for temporal dependencies nor the seasonality patterns that are ubiquitous in MTS, therefore, their performance is often lacking.

\noindent Deep learning models have garnered much attention in recent years, and there have been two main types of algorithms used in AD domain. One is autoencoder based \cite{autoencoder}. For example, \cite{autoencoder1} investigated the use of Gaussian classifiers with auto-encoders to model density distributions in multi-dimensional data. \cite{mscred} proposed a convolutional LSTM encoder-decoder structure to capture the temporal dependencies in time series, while assigning root causes for the anomalies. These models achieved better performance compared to the classical machine learning models. However, they do not model seasonality patterns, and they are built under the assumption that the training data do not contain contamination. Furthermore, they did not fully explore the power of a discriminator and a generator which has shown to have a superior performance in computer vision domain. This leads to the other type of deep learning algorithms: generative adversarial networks (GANs). Several recent studies demonstrated that the use of GANs has great promise to detect anomalies in images by mapping high-dimensional images to low dimensional latent space \cite{ganomaly,gan3image,gan4image}. However, these models have an unrealistic assumption that the training data is contamination free. A weak labeling to inspect this condition would make such algorithms not being fully unsuprvised. Further, applying GANs to data structures other than images is challenging and under-explored. To the best of our knowledge, we are among the first to extend the applications of GANs to MTS.

\section{Methodology}\label{method}
We define an MTS $X=(X_1^T,...,X_n^T)\in\mathbb{R}^{n\times T}$, where $n$ is the number of time series, and $T$ is the length of the historical training data. 
We aim to predict two AD outcomes: 1) the time points $t$ after $T$ that are associated with anomalies, and 2) time series $x_i, i \in \{1,..,n\}$ causing the anomalies. 
In the following section, we first describe how we reconstruct the input MTS to be consumed by a convolutional GAN. Then we introduce the RSM-GAN framework by decomposing it into three components: the architecture, the inner-structure, and the attention mechanism for seasonality adjustment. Finally, after the model is trained, we describe how we develop an anomaly scoring and casual inference procedure to identify anomalies and the root causes in the test data.


\subsection{RSM-GAN Framework}
\subsubsection{MTS to Image Conversion} To extend GAN to MTS and to capture inter-correlation between multiple time series, we convert the MTS into an image-like structure through the construction of the so-called multi-channel correlation matrix (MCM), inspired by \cite{song2018deep,mscred}. 
Specifically, we define multiple windows of different sizes $W=(w_1,...,w_c)$, and calculate the pairwise inner product (correlation) of time series within each window. For a specific time point $t$, we generate $c$ matrices (channels) of shape $n\times n$, where each element of matrix $S_t^w$ for a window of size $w$ is calculated by this formula:
\begin{equation}
    s_{ij}=\frac{\sum_{\delta=0}^{w}x_i^{t-\delta}\cdot x_j^{t-\delta}}{w}
\end{equation}

\noindent In this work, we select windows $W=(5, 10, 30)$. This results in $3$ channels of $n\times n$ correlation matrices. To convert the span of MTS into this shape, we consider a step size $s_s=5$. Therefore, $X$ is transformed to $S=(S_1,...,S_M)\in\mathbb{R}^{M\times c \times n\times n}$, where  $M=\lfloor\frac{T}{s_s}\rfloor$ steps presented by MCMs. Finally, to capture the temporal dependency between consecutive steps, we stack $h=4$ previous steps to the current step $t$ to prepare the input to the GAN-based model. Later, we extend MCM to also capture seasonality unique to MTS.

\subsubsection{RSM-GAN Architecture} 
\begin{figure}
  \centering 
  \includegraphics[width=0.88\columnwidth]{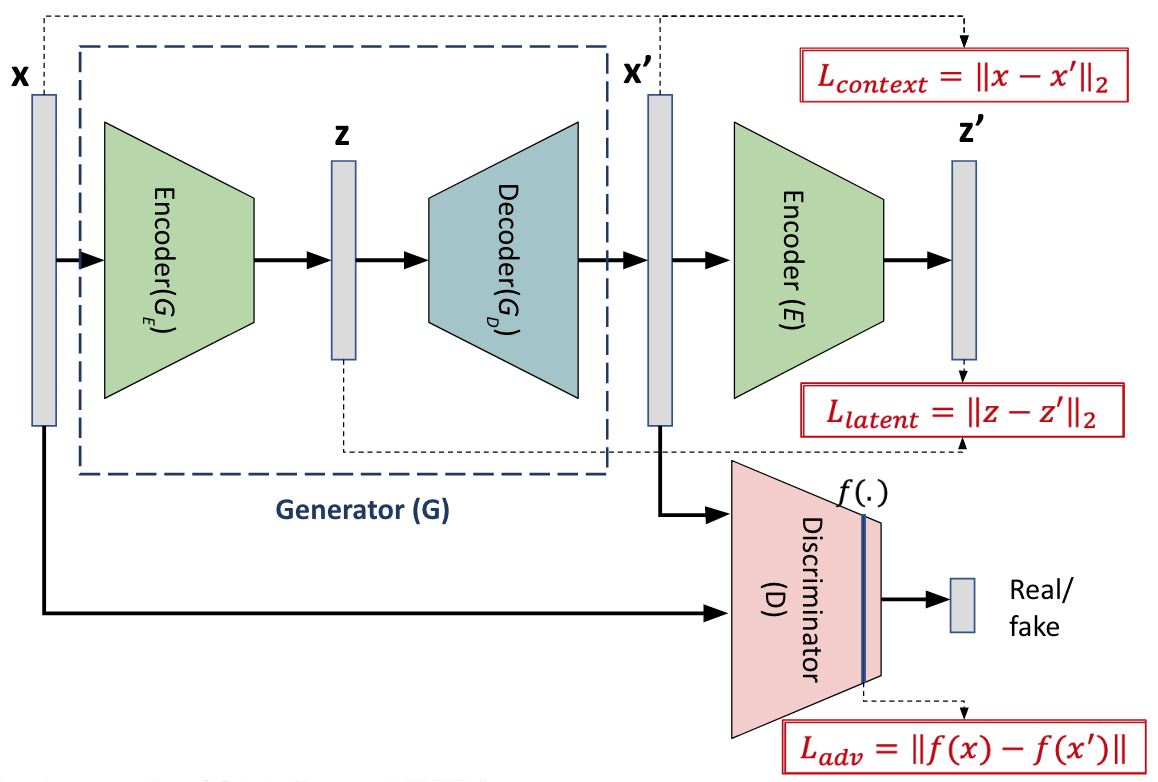}
  \caption{GAN architecture with loss definitions}
  \label{fig:GAN} 
\end{figure}

The idea behind using a GAN to detect anomalies is intuitive. During training, a GAN learns the distribution of the input data. Then, if anomalies are present during testing, the networks would fail to reconstruct the input, thus produce large losses. A GAN also exploits the power of the discriminator to optimize the network more efficiently towards training the distribution of input. However, in most GAN literature the training data is explicitly assumed to be normal with no contamination. \cite{encoder} have shown in a study that simultaneous training of an encoder with GAN improves the robustness of the model against contamination. To this end, we adopt an encoder-decoder-encoder structure \cite{ganomaly}, with the additional encoder, to capture the training data distribution in both original and latent space. It improves the robustness of the model to training noise, because the joint encoder forces similar inputs to lie close to each other also in the latent space.  
Specifically, in Figure \ref{fig:GAN} the generator $G$ has autoencoder structure that the encoder ($G_E$) and decoder ($G_D$) interact with each other to minimize the reconstruction or contextual loss: the $l_2$ distance between input $x$ and reconstructed input $x'$. Furthermore, an additional encoder $E$ is trained jointly with the generator to minimize the latent loss: the $l_2$ distance between latent vector $z$ and reconstructed latent vector $z'$. Finally, the discriminator $D$ is tasked to distinguish between the original input $x$ and the generated input $G(x)$. Following the recent improvements on GAN-based image AD \cite{gan2,gan3image}, we use feature matching loss for the adversarial training. Feature matching exploits the internal representation of the input $x$ induced by an intermediate layer in $D$. Assuming that the function $f(\cdot)$ will produce such representation, the discriminator aims to maximize the distance between $f(x)$ and $f(x')$ to effectively distinguish between original and generated inputs. At the same time, the generator battles against the adversarial loss to confuse the discriminator. With multiple loss components and training objectives, we employ the Wasserstein GAN with gradient penalty (WGAN-GP) \cite{wgan-gp} to 1) enhance the training stability, and 2) converge faster and more optimally. The final objective functions for the generator and discriminator (critic) are as following:
\begin{equation}
    L_D = \max_{w \in W} \mathbb{E}_{x\sim p_x}[f_w(x)] - \mathbb{E}_{x\sim p_x} [f_w(G(x))]
\end{equation}

\begin{equation}
\begin{aligned}
    L_G = \min_{G}\min_{E}  \Big( & w_1\mathbb{E}_{x\sim p_x} \parallel x-G(x) \parallel_2 + \\
    & w_2 \mathbb{E}_{x\sim p_x} \parallel G_E(x)-E(G(x)) \parallel_2 + \\
    & w_3  \mathbb{E}_{x\sim p_x}[f_w(G(x))]\Big)
\end{aligned}
\end{equation}
Where $w_1$, $w_2$, and $w_3$ are weights controlling the effect of each loss on the total objective. We employ 
Adam optimizer \cite{adam} to optimize the above losses.

\noindent In the next section, we describe how we design the internal structure of each network in RSM-GAN to capture the spatial as well temporal dependencies in our input data.

\subsubsection{Internal Encoder and Decoder Structure} 
\begin{figure}
  \centering 
  \includegraphics[width=0.5\textwidth]{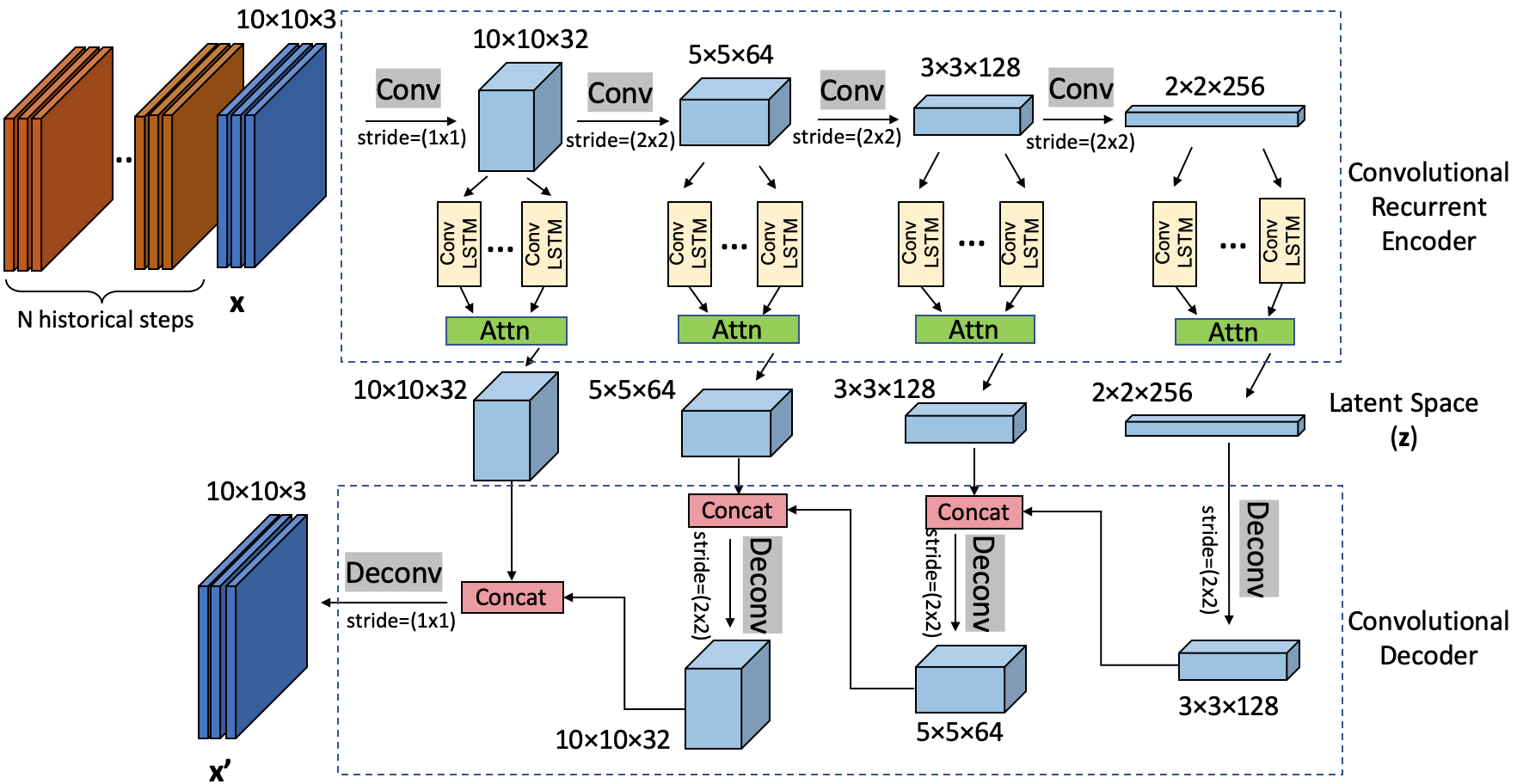} 
  \caption{Inner structure of convolutional recurrent encoders and convolutional decoder (with $n=10$) \cite{mscred}}
  \label{fig:ED_internal} 
\end{figure}

In addition to the convolutional layers in the encoders, we add RNN layers to jointly capture the spatial patterns and temporality of our MCM input by using convolutional-LSTM (convLSTM) gates. We apply convLSTM to every convolutional layer due to its optimal mapping to the latent space \cite{mscred}. 
The convolutional decoder applies multiple deconvolutional layers in reverse order to reconstruct MCM at current time step. Starting from the last convLSTM output, decoder applies deconvolutional layer and concatenates the output with convLSTM output of the previous step. The concatenation output is further an input to the next deconvolutional layer, and so on. Figure \ref{fig:ED_internal} illustrates the detailed inner-structure of the encoder and decoder networks.

\noindent The second encoder $E$ follows the same structure as the generator's encoder $G_E$ to reconstruct latent space $z'$. Input to the discriminator is the original or generated MCM of each time step. Therefore, internal structure of the discriminator consist of three simple convolutional layers, with the last layer representing $f(\cdot)$.

\subsubsection{Seasonality Adjustment via Attention Mechanism} 
\begin{figure}[ht!]
  \centering 
  \includegraphics[width=0.9\columnwidth]{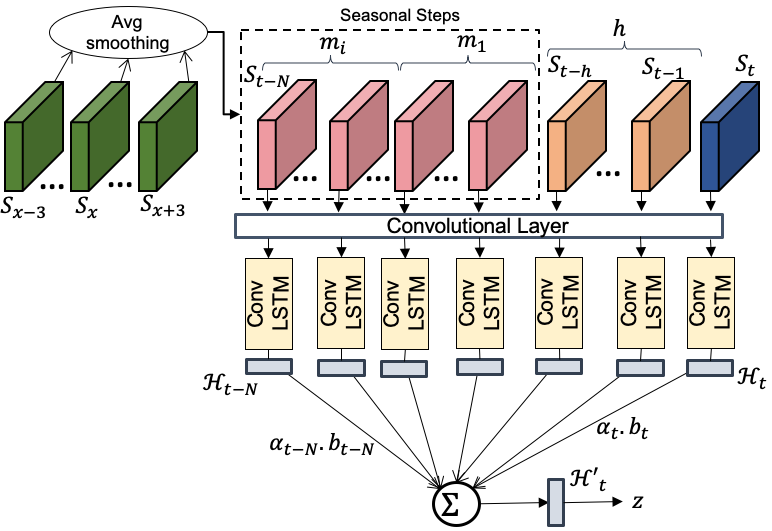} 
  \caption{Smoothed attention mechanism}
  \label{fig:attention} 
\end{figure}

The construction of the initial MCM does not take into consideration of the seasonality. We propose to first stack previous seasonal data points to the input data, and then let the convLSTM model temporal dependencies through attention mechanism. Specifically, in addition to $h$ previous immediate steps, we add $m_i$ previous seasonal steps per seasonal pattern $i$. To illustrate, assume the input has both the daily and weekly seasonality. For a certain time $t$, we stack MCMs of up to $m_1$ days ago at the same time, and up to $m_2$ weeks ago at the same time.  
Additionally, to account for the fact that seasonal patterns are often not exact, we smooth the seasonal steps by averaging over steps in a neighboring window of $30$ minutes.\\
Moreover, even though the $h$ previous steps are closer to the current time step, but the previous seasonal steps might be a better indicator to reconstruct the current step. Therefore, we further apply an attention mechanism to the convLSTM layers, and let the model decide the importance of all prior steps based on the similarity rather than recency. Attention weights are calculated based on the similarity of the hidden state representations in the last layer, by the following formula:
\begin{equation}
    \mathcal{H'}_t = \sum_{i\in (t-N,t)} \alpha_i \mathcal{H}_i,
    \alpha_i=\mathrm{softmax}\Big(\frac{Vec(\mathcal{H}_t)^T Vec(\mathcal{H}_i)}{\mathcal{X}}\Big)
\end{equation}
Where $N=h+\Sigma m_i$, $Vec(\cdot)$ denotes the vector, and $\mathcal{X}=5$ is the rescaling factor. Figure \ref{fig:attention} presents the structure of the described smoothed attention mechanism.
Finally, to make our model even more adaptable to real-world datasets that often exhibit holiday effects, we multiply the attention weight $\alpha_i$ by a binary bit $b_i \in \{0,1\}$, where $b_i=0$ in case of holidays or other exceptional behavior in previous steps. This way, we eliminate the effect of undesired steps from the current step.

\subsection{Testing Phase}
\subsubsection{Anomaly Score Assignment} After training the RSM-GAN, 
an anomaly score is assigned based on the residual MCM of the first channel (context) and latent vector of the RSM-GAN. We define broken tiles as the elements of the contextual or latent residual matrix that have error value of greater than $\theta_b$. \cite{mscred} defined a scoring method based on the number of broken tiles in contextual or latent residual matrix (context$_{b}$ and latent$_{b}$). However, this method is more sensitive to severe anomalies, and lowering the threshold results in high FPR. We propose a root cause-based counting procedure. Since each row/column in the contextual residual matrix is associated with a time series, the ones with the largest number of broken tiles are contributing the most to the anomalies. Therefore, by defining a threshold $\theta_h \leq \theta_b$, we only count the number of broken tiles in rows/columns with more than half broken. We name our new scoring method context$_{h}$.
The above thresholds $\theta = \beta \times \eta_{.996}(E_{train})$, which is calculated based on $99.6^{th}$ percentile of error in the training residual matrices, and the best $\beta$ is captured by grid search on the validation set. 
\subsubsection{Root Cause Framework} Large errors associated with elements of rows/columns of the residual MCM are indicative of anomalous behavior in those time series. To identify those abnormal time series, we need a root-cause scoring system to assign a score to each time series based on severity of its associating errors. We present $3$ different methods: 1) number of broken tiles (using the optimized $\theta$ from previous step), 2)the weighted sum of broken tiles based on their absolute error, and 3) the sum of absolute errors. 

\noindent Furthermore, the number of root causes, $k$, is unknown in real-life applications. \cite{mscred} used an arbitrary number of 3. Here, we propose to use an elbow method \cite{elbow} to find the optimal $k$ number of time series from the root cause scores. In this approach, by sorting the scores and plotting the curve, we aim to find the point where the amount of errors become very small and close to each other. Basically, for each point $n_i$ on the score curve, we find the point with maximum distance from a vector that connects the first and last scores. Time series associated with the scores greater than this point are identified as root causes.

\section{Experimental Setup}
\subsection{Data}
To evaluate different aspects of RSM-GAN, we conduct a comprehensive set of experiments by generating synthetic time series with multiple settings, along with a real-world encryption key dataset.
\subsubsection{Synthetic Data}
To simulate data with different seasonality and contamination, we first generate sinusoidal-based waves of length $T$ and periodicity $F$:
\begin{equation}
    S(t, F) = \left\{
    \begin{array}{ll}
        \sin[(t-t_0)/F]+0.3\times\epsilon_t & s_{rand}=0 \\
        \cos[(t-t_0)/F]+0.3\times\epsilon_t & s_{rand}=1
    \end{array}
\right.
\end{equation}
Where $s_{rand}$ is 0 or 1, $t_0 \in [10,100]$ is shift in phase and they are randomly selected for each time series. $\epsilon_t \sim \mathcal{N}(0,1)$ is the random noise, and $F \in [60,100]$ is the periodicity or seasonality. Ten time series with 2 months worth of data by minute sampling frequency are generated, or $T=80,640$. Each time series with combined seasonality is generated by:
\begin{equation}
    S(t) = S(t, F_{rand})+ S(t, F_{day}) + S(t,F_{week})
\end{equation}
Where $F_{day} = \frac{2\pi}{60\times24}$ and $F_{week} = \frac{2\pi}{60\times24\times7}$. To simulate anomalies with varying duration and intensity, we shock time series with a random duration ($[5,60]$ minutes), direction, and number of root causes ($[2,6]$). Each experiment is conducted with different seasonality patterns and contamination settings.
\subsubsection{Encryption Key Data} Our encryption-key dataset contains $7$ time series generated from a project's encryption process. Each time series represents the number of requests for a specific encryption key per minute. The dataset contains $4$ months of data or $T=156,465$. Four anomalies with various length and scales are identified in the test sequence by a security expert, and we randomly injected $5$ additional anomalies into both the train and test sequences.
\subsection{Baseline Models}
Three baseline models are used for comparison. Two are classical machine learning models, i.e., One-class SVM (OC-SVM) \cite{one-svm} and Isolation Forest \cite{iforest}. We also compare our model performance with that of MSCRED \cite{mscred} with the same input as ours. MSCRED is run in a sufficient number of epochs and its best performance is reported.

\subsection{Evaluation Metrics}
In addition to precision, recall, false positive rate, and F1 score, we include the \textbf{Numenta Anomaly Benchmark (NAB)} score \cite{numenta}. NAB is a standard open source framework for evaluating real-time AD algorithms. The NAB assigns score to each positive detection based on their relative position to the anomaly window by a scaled \textit{sigmoid} function (between -1 and 1). Specifically, it assigns a positive score to the earliest detection within anomaly window (1 to the beginning of the window) and negative score to detections after the window (false positives). Additionally, it assigns a negative score (-1) to the missed anomalies.
NAB score is more comprehensive than standard metrics because 
it also rewards timely detection. Early detection is critical for high-stake AD tasks such as cyber-attack monitoring. Also, it penalizes false positives as they get farther from the true anomaly window due to high cost of the manual inspection of the system. 

\noindent In our experiments, the first half of the time series are used for training the model and the remainder for evaluation. RSM-GAN is implemented in Tensorflow and trained in 300 epochs, in batches of size $32$, on an AWS Sagemaker instance with four $16$GB GPUs. 
All the results on synthetic data are produced by an average over five runs.

\section{Result and Discussion}
\subsection{Anomaly Score Assignment}
We first evaluate our new score assignment method context$_h$ against the $3$ other methods described before. 
Table \ref{tab:scores} reports the performance of RSM-GAN on synthetic MTS with no contamination and seasonality, using different scoring methods. The reported threshold is the optimum threshold obtained by the grid search. As we can see, our proposed context$_{h}$ method results in more precise predictions and has the highest NAB score. Specifically,  context$_{h}$ improves the precision and FPR by $6.2\%$ and $0.08\%$ compared to the context$_{b}$ method. Scoring based on the latent residual loss results in the lowest performance. Also, combining the methods by calculating a weighted sum of context and latent-based scores does not help improving the performance. Further, this performance comparison maintains the same for other more complex synthetic settings. Thus, \textbf{context$_{h}$} will be the scoring method reported in the subsequent sections. 

\begin{table}[htb]
\caption{Model performance with different anomaly score assignment methods}
\begin{adjustbox}{width=0.96\columnwidth,center}
\begin{tabular}{c|c|ccccc}
\textbf{Score} &\textbf{Threshold}&\textbf{Precision}& \textbf{Recall} & \textbf{F1} &\textbf{FPR} & \textbf{NAB Score} \\ \hline
latent$_{b}$ & 0.0099 & 0.648 & 0.819 &	0.723 &	0.0040 &	0.460 \\
context$_{b}$ & 0.0019 & 0.784 & \textbf{0.958} &	0.862 &	0.0023 &	0.813 \\
context$_{h}$ & 0.00026 &\textbf{0.846} &	0.916 &	\textbf{0.880} &	\textbf{0.0015} &	\textbf{0.859} \\
combined & - & 0.767 &	0.916 &	0.835 &	0.0025 &	0.721 \\
\end{tabular}
\end{adjustbox}
\label{tab:scores}
\end{table}

\subsection{Root Cause Identification Assessment}
RSM-GAN detects root causes using context-based residual matrix. In this section, we compare the results of MSCRED and RSM-GAN using $3$ root cause scoring methods. Root causes are identified based on the average of errors per time series in an anomaly window and by applying the aforementioned elbow-based identification method. Precision, recall and F1 scores are averaged over all detected anomalies. 

\begin{table}[ht]
\caption{Root cause identification performance with different root cause scoring methods }
\begin{adjustbox}{width=0.9\columnwidth,center}
\begin{tabular}{c||c|ccc}
\textbf{Model} &\textbf{Scoring}&\textbf{Precision}& \textbf{Recall} & \textbf{F1} \\ \hline
\multirowcell{3}{MSCRED} 
& Number of broken (NB) & \textbf{0.5154} &	\textbf{0.7933} &	\textbf{0.6249} \\
& Weighted broken (WB) & 0.5071 &	0.6866 &	0.5834 \\
& Absolute error (AE) & 0.5504 &	0.7066 &	0.6188 \\ \hline

\multirowcell{3}{RSM-GAN} 
& Number of broken (NB) & 0.4960 &	0.8500 &	0.6264 \\
& Weighted broken (WB) & \textbf{0.6883} &	\textbf{0.8666} &	\textbf{0.7672} \\
& Absolute error (AE) & \textbf{0.6883} &	\textbf{0.8666} &	\textbf{0.7672} \\ \hline

\end{tabular}
\end{adjustbox}
\label{tab:rootcause}
\end{table}

\begin{table*}[ht]
\caption{Model Performance with different levels of training data contamination}
\begin{adjustbox}{width=0.73\textwidth,center}
\begin{tabular}{c||c|ccccc|c}
\textbf{Contamination} &\textbf{Model}&\textbf{Precision}& \textbf{Recall} & \textbf{F1} &\textbf{FPR} & \textbf{NAB Score} & \textbf{Root Cause Recall} \\ \hline
\multirowcell{4}{No contamination \\ train: 0 (0) \\ test: 10 (\%0.82)} & OC-SVM & 0.1581 &	\textbf{1.0000} &	0.2730 &	0.0473 &	-8.4370 & - \\
& Isolation Forest & 0.0326 &	\textbf{1.0000} &	0.0631 &	0.2640 &	-51.4998 & - \\
& MSCRED & 0.8000 &	0.8450 &	0.8219 &	0.0018 &	0.7495 & \textbf{0.7533} \\
& RSM-GAN & \textbf{0.8461} &	0.9166 &	\textbf{0.8800} &	\textbf{0.0015} &	\textbf{0.8598} &	0.6333 \\ \hline

\multirowcell{4}{Mild contamination \\ train: 5 (\%0.43) \\ test: 10 (\%0.76)} 
& OC-SVM & 0.2810 &	\textbf{1.0000} &	0.4387 &	0.0218 &	-3.3411 & - \\
& Isolation Forest & 0.3134 &	\textbf{1.0000} &	0.4772 &	0.0187 &	-2.7199 & - \\
& MSCRED & 0.6949 &	0.6029 &	0.6457 &	0.0023 &	0.2721 &	0.5483 \\
& RSM-GAN & \textbf{0.8906} &	0.7500 &	\textbf{0.8143} &	\textbf{0.0009} &	\textbf{0.8865} &	\textbf{0.7700} \\ \hline

\multirowcell{4}{Medium contamination \\ train: 10 (\%0.82) \\ test: 10 (\%0.85)} 
& OC-SVM & 0.4611 &	\textbf{1.0000} &	0.6311 &	0.0113 &	-1.2351 & - \\
& Isolation Forest & 0.6311 &	\textbf{1.0000} &	0.7739 &	0.0056 &	-0.1250 & - \\
& MSCRED & 0.6548 &	0.7143 &	0.6832 &	0.0036 &	0.2712 &	0.6217 \\
& RSM-GAN & \textbf{0.8553} &	0.8442 &	\textbf{0.8497} &	\textbf{0.0014} &	\textbf{0.8511} &	\textbf{0.8083} \\ \hline

\multirowcell{4}{Severe contamination \\ train: 15 (\%1.19) \\ test: 15 (\%1.18)} 
& OC-SVM & 0.5691 &\textbf{	1.0000} &	0.7254 &	0.0102 &	-0.3365 & - \\
& Isolation Forest & 0.8425 &	\textbf{1.0000} &	\textbf{0.9145} &	0.0025 &	0.6667 & - \\
& MSCRED & 0.5493 &	0.7290 &	0.6265 &	0.0080 &	0.0202 &	0.6611 \\
& RSM-GAN & \textbf{0.8692} &	0.8774 &	0.8732 &	\textbf{0.0018} &	\textbf{0.8872} &	\textbf{0.8133} \\ \hline

\end{tabular}
\end{adjustbox}
\label{tab:contamination}
\end{table*}

\noindent The synthetic data used in this experiment has two combined seasonal patterns and ten anomalies in the training data. Table \ref{tab:rootcause} shows root cause identification performance of RSM-GAN and MSCRED.  Overall, RSM-GAN outperforms MSCRED. As the results suggest, the NB method performs the best for MSCRED. However, for RSM-GAN the WB and AE methods leads to the best performance. Since the same result holds for other settings, we report NB for MSCRED and AE for RSM-GAN in subsequent sections.

\subsection{Contamination Tolerance Assessment}
In this section, we assess the robustness of RSM-GAN with different levels of contamination in training data, and the results are compared to the baseline models. In this experiment, 
the level of contamination starts with no contamination and at each subsequent level, we add $5$ more random anomalies with varying duration to the training data. The percentages presented in the Contamination column in Table \ref{tab:contamination} shows the proportions of the anomalous time points in train/test time span. 

\noindent Results in Table \ref{tab:contamination} suggest that our proposed model outperforms all baseline models at all contamination levels for all metrics except of the recall. Note that the \%100 recall for classic baseline models is at the expense of FPR as high as $26.4\%$, especially for the less severe contamination. Furthermore, comparison of the NAB scores shows that our model has more timely detections and false positives are within a window of the anomalies.

\noindent Lastly, as we can see, the MSCRED performance drops drastically as the contamination level increases. This is because the encoder-decoder structure of this model cannot handle high levels of contamination while training.

\subsection{Seasonality Adjustment Assessment}
\noindent In this section, we assess the performance of our proposed attention mechanism for capturing seasonality in MTS. In many of the real-world AD applications, time series might contain a single or multiple seasonal patterns (daily/weekly/monthly/etc.), with the effect of special events, like holidays. The performance of RSM-GAN is assessed in different seasonality settings. In the first three experiments,  synthetic MTS (2 months, sampled per minute) are generated with no training data contamination and no seasonality, then daily and weekly seasonality patterns are added one by one. In the last experiment, we simulate $3$ years of hourly data, and add special patterns for the time steps related to the US holidays in both the train and test sets. The test set of each experiment is contaminated with $10$ random anomalies.

\noindent Comparing the results in Table \ref{tab:seasonality}, RSM-GAN shows consistent performance thanks to the attention mechanism capturing the seasonality patterns. All the other baseline models, especially MSCRED's performance deteriorated with increased complexity of the seasonal patterns. Precision is the main metric that drops drastically for the baseline models as we add more seasonality. This is because they do not account for changes due to seasonality and identify them as anomalies, which also led to high FPR.

\begin{table*}[tb!]
\caption{Model performance on synthetic data with different seasonal patterns and no contamination}
\begin{adjustbox}{width=0.73\textwidth,center}
\begin{tabular}{c||c|ccccc|c}
\textbf{Seasonality} &\textbf{Model}&\textbf{Precision}& \textbf{Recall} & \textbf{F1} &\textbf{FPR} & \textbf{NAB Score} & \textbf{Root Cause Recall} \\ \hline

\multirowcell{4}{Random seasonality} 
& OC-SVM & 0.4579 &	0.9819 &	0.6245 &	0.0097 &	-8.6320 & - \\
& Isolation Forest & 0.0325 &	\textbf{1.0000} &	0.0630 &	0.2646 &	-51.606 &- \\
& MSCRED & 0.8000 &	0.8451 &	0.8219 &	0.0019 &	0.7495 & \textbf{0.7533} \\
& RSM-GAN & \textbf{0.8462} &	0.9167 &	\textbf{0.8800} &	\textbf{0.0015} &	\textbf{0.8598} & 0.6333 \\ \hline

\multirowcell{4}{Daily seasonality} 
& OC-SVM & 0.1770 &	\textbf{1.0000} &	0.3008 &	0.0532 &	-9.5465 &  - \\
& Isolation Forest & 0.1387 &	\textbf{1.0000} &	0.2436 &	0.0710 &	-13.107 & - \\
& MSCRED & 0.7347 &	0.7912 &	0.7619 &	0.0033 &	0.3775 & \textbf{0.7467} \\
& RSM-GAN & \textbf{0.9012} &	0.7935 &	\textbf{0.8439} &	\textbf{0.0010} &	\textbf{0.5175} & 0.6717 \\ \hline

\multirowcell{4}{Daily and weekly \\ seasonality} 
& OC-SVM & 0.1883 &	\textbf{0.9487} &	0.3142 &	0.0400 &	-6.9745 &  - \\
& Isolation Forest & 0.1783 &	\textbf{0.9487} &	0.3002 &	0.0428 &	-7.5278 &  - \\
& MSCRED & 0.6548 &	0.7143 &	0.6832 &	0.0036 &	0.2712 &	\textbf{0.6217} \\
& RSM-GAN & \textbf{0.9000} &	0.6750 &	\textbf{0.7714} &	\textbf{0.0008} &	\textbf{0.5461} & 0.4650 \\ \hline

\multirowcell{4}{Weekly and monthly \\ seasonality \\ with holidays} 
& OC-SVM & 0.2361 &	\textbf{0.9444} &	0.3778 &	0.0425 &	-1.7362 & - \\
& Isolation Forest & 0.2783 &	0.8889 &	0.4238 &	0.0321 &	-1.0773 &  - \\
& MSCRED & 0.0860 &	0.7059 &	0.1534 &	0.0983 &	-5.1340 & 0.6067 \\
& RSM-GAN & \textbf{0.6522} &	0.8108 &	\textbf{0.7229} &	\textbf{0.0063} &	\textbf{0.5617} & \textbf{0.8667} \\ \hline
\end{tabular}
\end{adjustbox}
\label{tab:seasonality}
\end{table*}

\begin{figure}
  \centering 
  \includegraphics[width=\columnwidth]{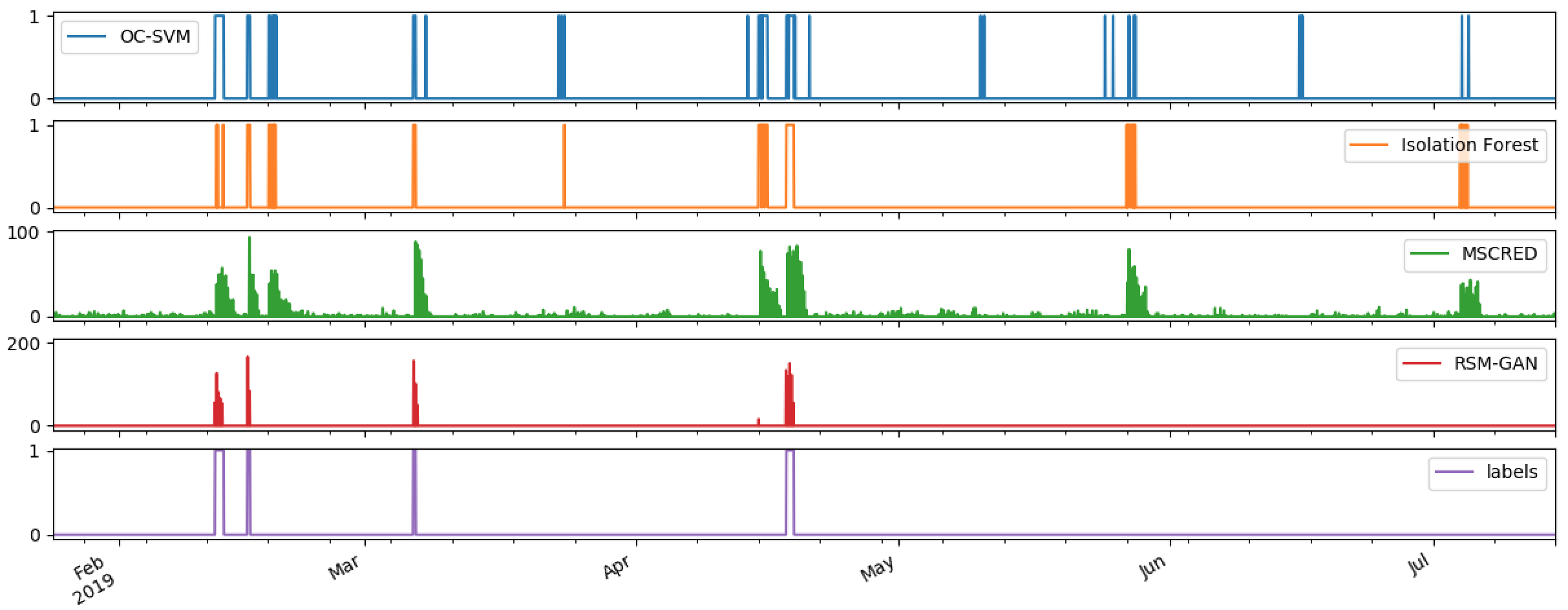} 
  \caption{Performance comparison on synthetic data with weekly and monthly seasonality and holiday effect}
  \label{fig:weekmo-holiday} 
\end{figure}

\noindent In the last experiment in Table \ref{tab:seasonality}, all of the abnormalities injected to holidays are wrongly flagged by the baseline models as anomalies, since no holiday adjustment is incorporated in these models. This resulted in low precision and high FPR for those models. In RSM-GAN, multiplying the binary vectors of holidays with the attention weights enables it to account for the holidays, which leads to the best performance in almost all metrics. Figure \ref{fig:weekmo-holiday} shows the ground truth anomaly labels (bottom), and the anomaly scores assigned by each model to each time step while testing. It is evident that our model accurately accounts for the holidays (18 Feb, 19 May, 4 Jul) and has much lower FPR. 

\begin{table*}[t!]
\caption{Model performance on encryption key and synthetic seasonal MTS with contamination}
\begin{adjustbox}{width=0.7\textwidth,center}
\begin{tabular}{c||c|ccccc|c}
\textbf{Dataset} &\textbf{Model}&\textbf{Precision}& \textbf{Recall} & \textbf{F1} &\textbf{FPR} & \textbf{NAB Score} & \textbf{Root Cause Recall} \\ \hline

\multirowcell{4}{Encryption \\ key} 
& OC-SVM & 0.1532 &	0.2977 &	0.2023 &	0.0063 &	-17.4715& - \\
& Isolation Forest & 0.3861 &	\textbf{0.4649} &	0.4219 &	0.0028 &	-6.9343	& - \\
& MSCRED & 0.1963 &	0.2442 &	0.2176 &	0.0055 &	-1.1047	& 0.4709 \\
& RSM-GAN & \textbf{0.6852} &	0.4405 &	\textbf{0.5362} &	\textbf{0.0011} &	\textbf{0.2992}	& \textbf{0.5093} \\ \hline

\multirowcell{4}{Synthetic} 
& OC-SVM & 0.6772 &	0.9185 &	0.7772 &	0.0038 &	-2.7621 & - \\
& Isolation Forest & 0.7293 &	\textbf{0.9610} &	0.8221 &	0.0033 &	-2.2490 & - \\
& MSCRED & 0.6228 &	0.7403 &	0.6746 &	0.0043 &	0.2753 &	0.6600 \\
& RSM-GAN & \textbf{0.8884} &	0.8438 &	\textbf{0.8649} &	\textbf{0.0010} &	\textbf{0.8986} &	\textbf{0.7870} \\ \hline

\end{tabular}
\end{adjustbox}
\label{tab:realworld}
\end{table*}

\begin{figure}[hb!]
\centering
\begin{subfigure}[b]{\columnwidth}
    \centering
   \includegraphics[width=\columnwidth]{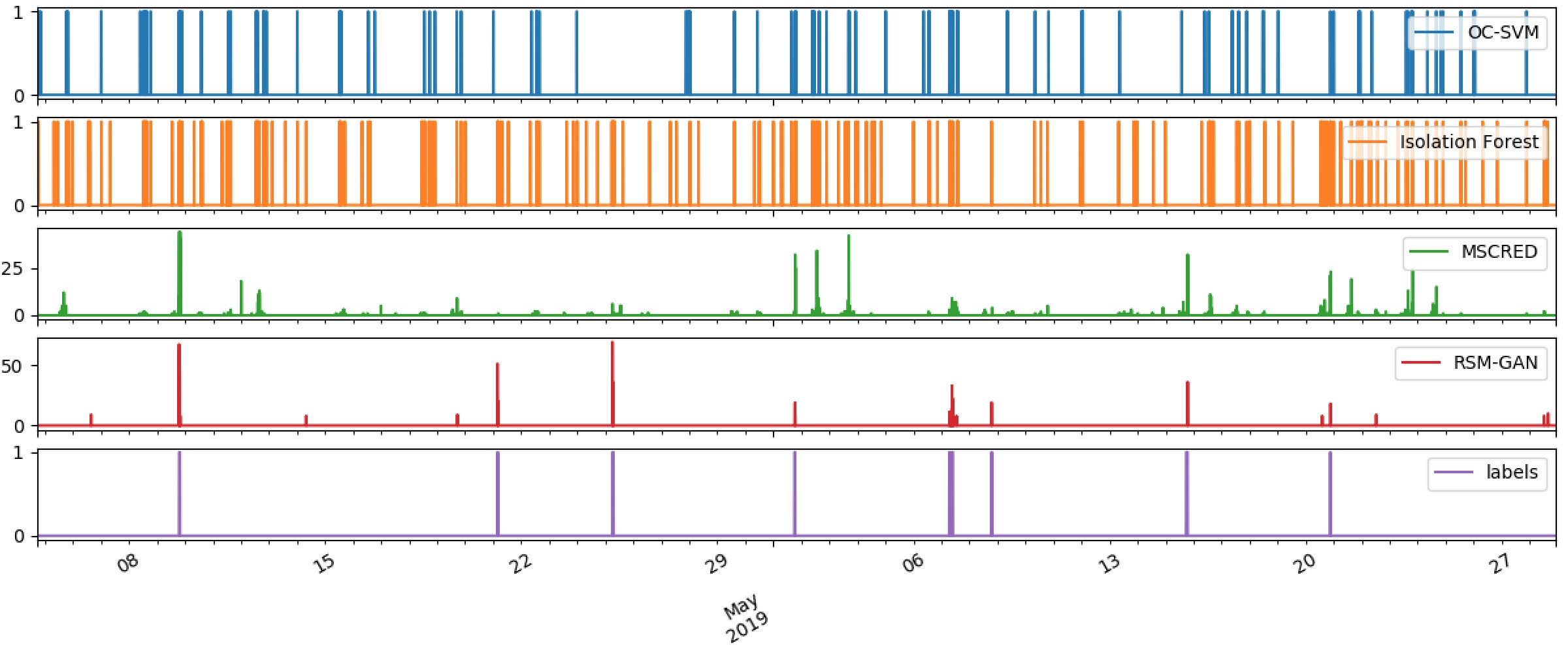}
   \caption{Encryption Key Dataset}
   \label{fig:final_real} 
\end{subfigure}

\begin{subfigure}[h!]{\columnwidth}
    \centering
   \includegraphics[width=\columnwidth]{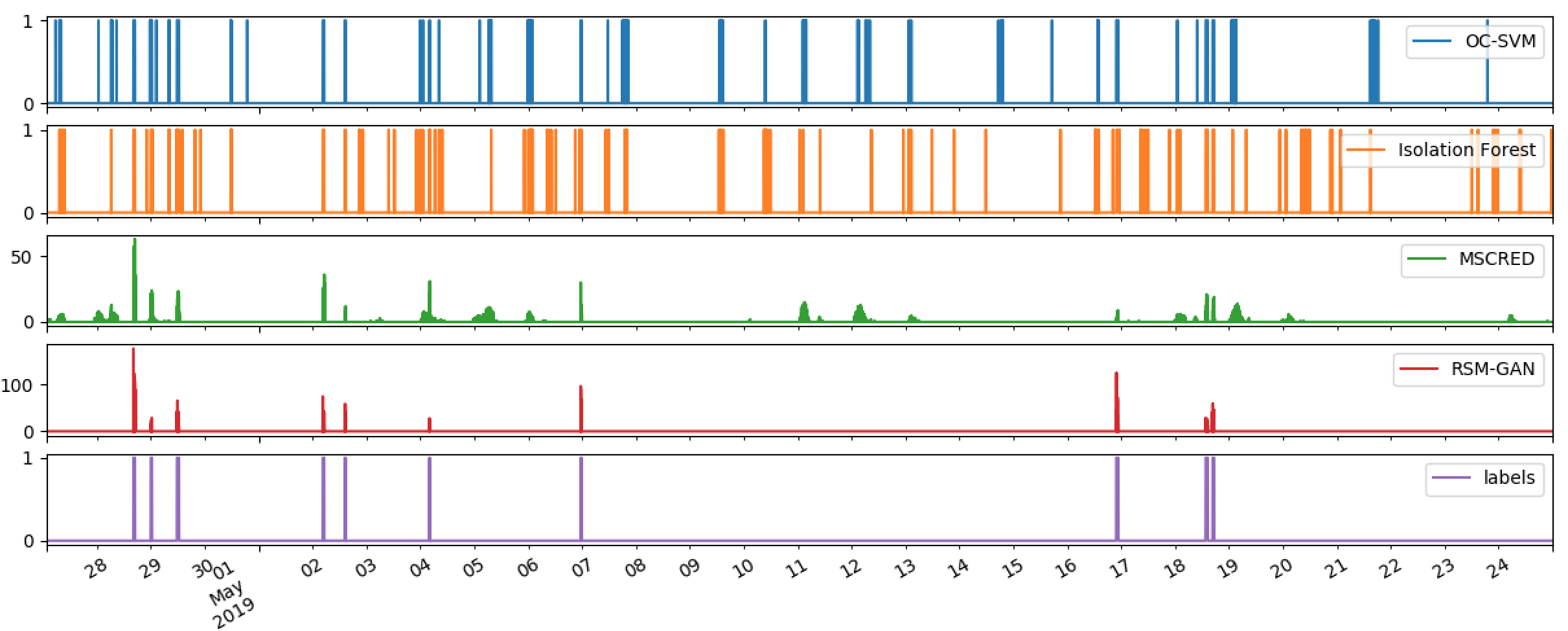}
   \caption{Synthetic Dataset}
   \label{fig:final_synth}
\end{subfigure}
\caption{Anomaly score assignment of different algorithms. The bottom plot is the ground truth labels.}
\label{fig:final_plot}
\end{figure}

\subsection{Performance on Real-world dataset}
\noindent This section evaluates our model on a real-world encryption key dataset. A cursory look shows that this dataset is noisy, and contains both daily and weekly seasonality.
To be comprehensive, we also create a synthetic dataset with similar patterns, i.e., daily and weekly seasonality as well as medium contamination (10 anomalies) in the training set. 
\noindent From Table \ref{tab:realworld}, we make the following observations: 1) RSM-GAN consistently outperforms all the baseline models in terms of detection and root cause identification recall for both the encryption key and the synthetic dataset. 2) Not surprisingly, for all the models, performance on the synthetic data is better than that of encryption key data. It is due to the excessive irregularities and noise in the encryption key data, and errors arising from ground truth labeling by experts. 3) The plots in Figure \ref{fig:final_plot} illustrates the anomaly scores assigned to each time point in test dataset by each algorithm. As we can see, even though isolation forest has the highest recall rate, it also detects many false positives not related to the actual anomaly windows, leading to negative NAB scores. As mentioned before, irrelevant false positives are costly in real-world applications. 4) By comparing our model to MSCRED in Figure \ref{fig:final_real} and Figure \ref{fig:final_synth}, we can see that MSCRED not only has much higher FPR, but it also fails to capture some anomalies. We conjecture it is because MSCRED's encoder-decoder structure is not as robust to the training data contamination, nor does it model the seasonality patterns.

\section{Conclusion}
In this work, we presented the challenges in MTS anomaly detection and proposed a novel GAN-based MTS anomaly detection framework (RSM-GAN) to solve those challenges. RSM-GAN takes advantage of the adversarial learning to accurately capture the temporal and spatial dependencies in the data, while simultaneously deploying an additional encoder to handle even severe levels of training data contamination. The novel attention mechanism in the recurrent layer of RSM-GAN enables the model to handle complex seasonal patterns often found in the real-world data. 
Furthermore, training stability and optimal convergence of the GAN is attained through the use of Wasserstein GAN with gradient penalty. We conducted extensive empirical studies and results show that our architecture together with a new score and causal inference framework lead to 
an exceptional performance over state-of-the-art baseline models on both synthetic and real-world datasets.

\bibliography{main}
\fontsize{9.8pt}{10.8pt}\selectfont
\bibliographystyle{aaai}

\end{document}